\documentclass[runningheads]{llncs}
\usepackage{graphicx}

\usepackage{tikz}
\usepackage{comment}
\usepackage{amsmath,amssymb} 
\usepackage{color}
\usepackage{url}

\usepackage[accsupp]{axessibility}  

\usepackage{multirow} 
\usepackage{makecell}
\usepackage{bm}
\usepackage{booktabs}
\usepackage{hyperref}
\DeclareFixedFont{\B}{OT1}{ptm}{bx}{n}{8pt}
\DeclareFixedFont{\BB}{OT1}{ptm}{bx}{n}{10pt}

\begin{document}
\pagestyle{headings}
\mainmatter
\def\ECCVSubNumber{1620}  

\title{CLIFF: Carrying Location Information in Full Frames into Human Pose and Shape Estimation} 

\titlerunning{CLIFF}

\author{
Zhihao Li \and
Jianzhuang Liu \and
Zhensong Zhang \and
Songcen Xu \and
Youliang Yan \thanks{Corresponding author.}
}

\authorrunning{Z. Li et al.}

\institute{
Huawei Noah's Ark Lab
\email{\{zhihao.li,liu.jianzhuang,zhangzhensong,xusongcen,yanyouliang\}@huawei.com}}
\maketitle

\begin{abstract}
Top-down methods dominate the field of 3D human pose and shape estimation, because they are decoupled from human detection and allow researchers to focus on the core problem. However, cropping, their first step, discards the location information from the very beginning, which makes themselves unable to accurately predict the global rotation in the original camera coordinate system. To address this problem, we propose to Carry Location Information in Full Frames (CLIFF) into this task. Specifically, we feed more holistic features to CLIFF by concatenating the cropped-image feature with its bounding box information. We calculate the 2D reprojection loss with a broader view of the full frame, taking a projection process similar to that of the person projected in the image. Fed and supervised by global-location-aware information, CLIFF directly predicts the global rotation along with more accurate articulated poses. Besides, we propose a pseudo-ground-truth annotator based on CLIFF, which provides high-quality 3D annotations for in-the-wild 2D datasets and offers crucial full supervision for regression-based methods. Extensive experiments on popular benchmarks show that CLIFF outperforms prior arts by a significant margin, and reaches the first place on the AGORA leaderboard (the SMPL-Algorithms track). The code and data are available at \url{https://github.com/huawei-noah/noah-research/tree/master/CLIFF}.

\end{abstract}

\section{Introduction}
\label{sec:intro}

Given a single RGB image, 3D human pose and shape estimation aims to reconstruct human body meshes with the help of statistic models \cite{anguelov2005scape,loper2015smpl,osman2020star,xu2020ghum}.
It is a fundamentally under-constrained problem due to the depth ambiguity.
This problem attracts a lot of research because of its key role in many applications such as AR/VR, telepresence, and action analysis.

With the popular parametric human model SMPL \cite{loper2015smpl}, regression-based methods \cite{kanazawa2018end,kocabas2021pare,moon2020pose2pose} learn to predict the SMPL parameters from image features in a data-driven way, and obtain the meshes from these predictions through a linear function.
Like most tasks in computer vision, there are two approaches to do this: top-down \cite{kanazawa2018end,kolotouros2019learning,kocabas2020vibe,kocabas2021pare} and bottom-up \cite{sun2021monocular,zhang2021bmp}.
The former first detects humans, then crops the regions of interest, and processes each cropped image independently.
The latter takes a full image as input and gives the results for all individuals at once.
The top-down approach dominates this field currently, because it is decoupled from human detection, and has high recall and precision performances thanks to the mature detection technique \cite{redmon2018yolov3,ren2015faster,chen2019hybrid}.

\begin{figure}[t]
	\centering
	\includegraphics[width=\textwidth]{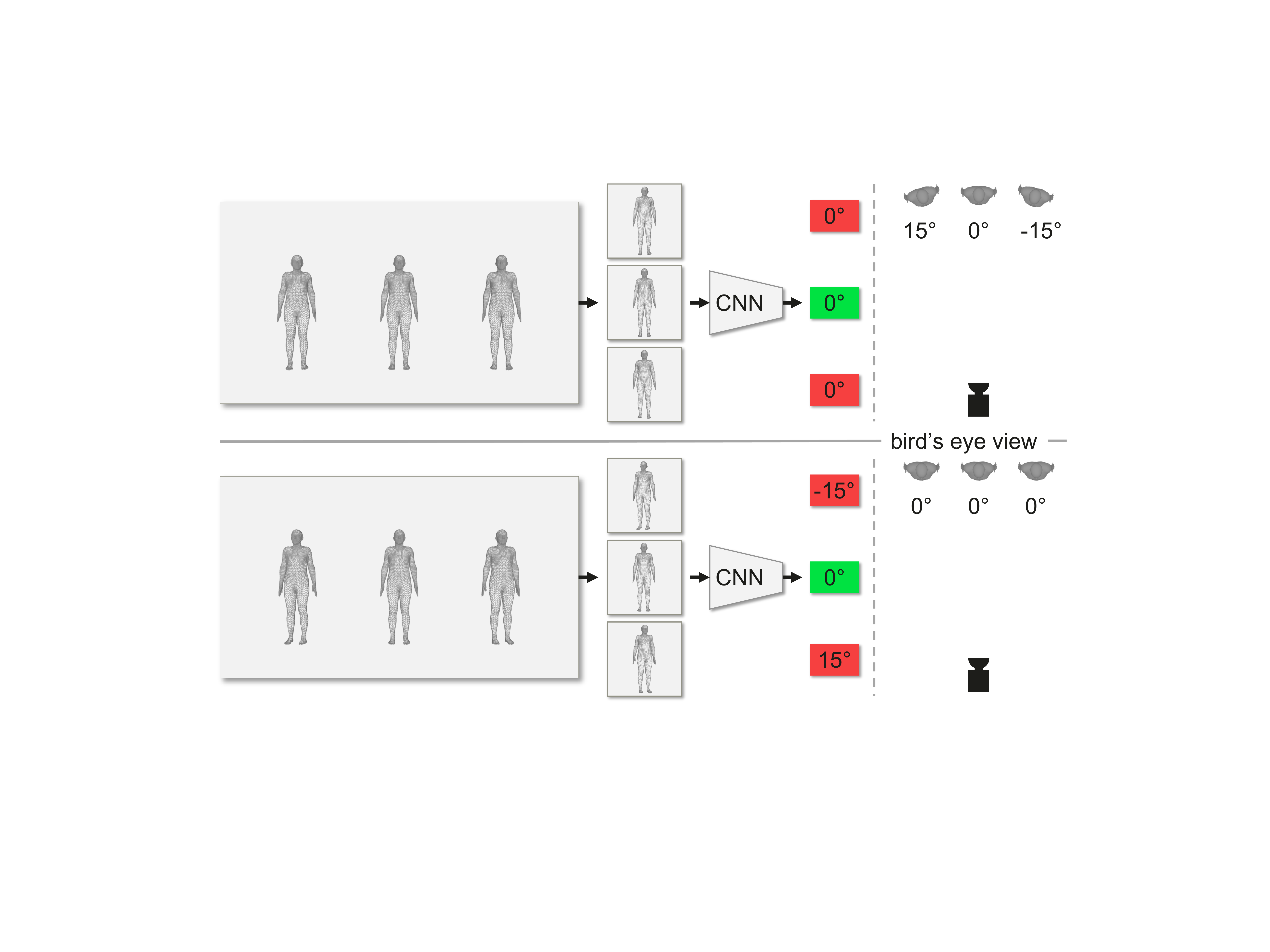}
	\caption{In the upper part, the cropped images look similar and thus get close predictions from a CNN model. However, two of the three predictions are wrong (marked in red). Actually, they have clearly different global rotations relative to the original camera, which can be seen from the bird's eye view.
	In the lower part, the three global rotations are the same, but again two estimations are wrong.
	We only show the yaw angles of the global rotations here for simplicity.}
	\label{fig:motivation}
\end{figure}

However, cropping, the first step in the top-down approach, discards location information from the very beginning, which is essential in estimating the global rotation in the original camera coordinate system with respect to the full image.
Take Fig.~\ref{fig:motivation} as an example, where the original images come from a perspective camera with a common diagonal Field-of-View 55°.
After cropping, the input images to the CNN model look similar, and thus get close predictions without surprise (see the upper part of Fig.~\ref{fig:motivation}).
In fact, the three persons have clearly different global rotations, which can be inferred from the full image.
The same problem exists for other 2D evidences such as 2D keypoints.
As a result, the 2D reprojection loss calculated in the cropped images is not a proper supervision, which tends to twist articulated poses to compensate the global rotation error.
In another word, missing the location information introduces extra ambiguity.

To fix this problem, we propose CLIFF: Carrying Location Information in Full Frames into 3D human pose and shape estimation, by making two major modifications to the previous top-down approach.
First, CLIFF takes more holistic features as input.
Besides the latent code of the resolution-fixed cropped image, the bounding box information is also fed to CLIFF, which encodes its discarded location and size in the original full image, providing the model with adequate information to estimate the global rotation.
Second, CLIFF calculates the 2D reprojection loss with a broader view of the full frame.
We obtain the predicted 3D joints in the original camera coordinate system, and project them onto the full image instead of the cropped one.
Consequently, these predicted 2D keypoints have a projection process and perspective distortion similar to those of the person projected in the image, which is important for them to correctly supervise 3D predictions in an indirect way.
Fed and supervised by global-location-aware information, CLIFF is able to predict the global rotation relative to the original camera along with more accurate articulated poses.


On the other hand, the regression-based methods need full supervision for SMPL parameters to boost their performances \cite{kolotouros2019learning}.
However, it costs a lot of time and effort to obtain these 3D annotations, using multi-view motion capture systems \cite{ionescu2013human3,mehta2017monocular} or a set of IMU devices \cite{trumble2017total,von2018recovering}.
Moreover, the lack of diversity in actors and scenes limits the generalization abilities of these data-driven models.
On the contrary, 2D keypoints are straightforward and inexpensive to annotate for a wide variety of in-the-wild images with diverse persons and backgrounds \cite{lin2014microsoft,andriluka20142d}.
Hence, some CNN-based pseudo-ground-truth (pseudo-GT) annotators \cite{kolotouros2019learning,joo2021exemplar,moon2020neuralannot} are introduced to lift these 2D keypoints up to 3D poses for full supervision.
Since these annotators are based on previous top-down models that are agnostic to the person location in the full frame, they produce inaccurate annotations, especially for the global rotation.

We propose a novel annotator based on CLIFF with a global perspective of the full frame, which produces high-quality annotations of human model parameters.
Specifically, we first pretrain the CLIFF annotator on several datasets with available 3D ground truth,
and then test it on the target dataset to predict SMPL parameters.
Using these predictions as regularization and ground-truth 2D keypoints as weak supervision, we finetune the pretrained model on the target dataset,
and finally test on it to infer SMPL parameters as pseudo-GT.
With the implicit prior of the pretrained weights and the explicit prior of the SMPL parameter predictions, the CLIFF annotator alleviates the inherent depth ambiguity to recover feasible 3D annotations from monocular images.


Our contributions are summarized as follows:
\begin{itemize}
\item We reveal that the global rotations cannot be accurately inferred when only using cropped images, which is ignored by previous methods, and propose CLIFF to deal with this problem by feeding and supervising the model with global-location-aware information.
\item Based on CLIFF, we propose a pseudo-GT annotator with strong priors to generate high-quality 3D annotations for in-the-wild images, which are demonstrated to be very helpful in performance boost.
\item We conduct extensive experiments on popular benchmarks and show that CLIFF outperforms prior arts by significant margins on several evaluation metrics (e.g., 5.7mm MPJPE and 6.5mm PVE on 3DPW), and reaches the first place on the AGORA leaderboard (the SMPL-Algorithms track).
\end{itemize}

\section{Related Work}

\subsubsection{3D human pose estimation.}
3D human pose is usually represented as a skeleton of 3D joints \cite{sharma2019monocular,wehrbein2021probabilistic,yu2021pcls,gong2021poseaug}, or a mesh of triangle vertices \cite{kanazawa2018end,kocabas2020vibe,moon2020i2l,choi2020pose2mesh,lin2021end,lin2021mesh}.
These vertex locations are inferred directly by model-free methods \cite{moon2020i2l,choi2020pose2mesh,lin2021end,lin2021mesh}, or obtained indirectly from parametric model (e.g., SMPL \cite{loper2015smpl}) predictions of model-based methods \cite{kanazawa2018end,kolotouros2019learning,kocabas2020vibe,sengupta2020synthetic,Rong_2019_ICCV}.
Optimization-based methods \cite{bogo2016keep,fang2021mirrored} are first proposed to iteratively fit the SMPL model to 2D evidences, while regression-based ones \cite{kanazawa2018end,pymaf2021} make the predictions in a straightforward way that may support real time applications.
Both top-down \cite{kolotouros2019learning,li2021hybrik,kocabas2021pare,lin2021mesh,zhang2020object} and bottom-up \cite{sun2021monocular,zhang2021bmp} approaches can do the job.
Because our method is a top-down framework to regress the SMPL parameters, we only review the most relevant work, and refer the reader to \cite{liu2021recent,tian2022hmrsurvey} for more comprehensive surveys.

\subsubsection{Input and supervision.}
Previous top-down methods take as input the cropped image \cite{kanazawa2018end} or/and 2D keypoints in the cropped region \cite{choi2020pose2mesh}, perhaps with additional camera parameters \cite{kocabas2021spec}.
Most of them project the predicted 3D joints onto the cropped image to compute the reprojection loss for supervision.
Since the location information is lost in the cropped image, it is difficult for them to estimate an accurate global rotation.
To solve this problem, Kissos et al. \cite{kissos2020beyond} use the prediction with respect to the cropped image as initialization, and then use SMPLify \cite{bogo2016keep} to refine the results for better pixel alignment.
Since SMPLify computes the reprojection loss in the full image, they obtain a better global rotation in the end.
However, as an optimization approach, SMPLify is very slow and may harm the articulated pose estimation \cite{joo2021exemplar,moon2020neuralannot}.
PCL \cite{yu2021pcls} warps the cropped image to remove the perspective distortion, and corrects the global rotation via a post-processing, but the articulated poses cannot be corrected.
In order to better estimate the root translation in multi-agent aerial applications, AirPose \cite{saini2022airpose} also provides the additional location information, but in a simpler way without clear geometric meanings.
CLIFF exploits the location information in both input and supervision, to predict more accurate global rotation and articulated poses simultaneously, without any post-processing.

\subsubsection{Pseudo-GT annotators.}
It tends to help regression-based models generalize better to train with diverse in-the-wild images.
However, it is hard to obtain the corresponding 3D ground truth, so pseudo-GT annotators are proposed.
Optimization-based annotators \cite{bogo2016keep} throw the images away, and fit the human model to 2D keypoints by minimizing the reprojection loss.
CNN-based annotators \cite{kolotouros2019learning,joo2021exemplar,moon2020neuralannot,kolotouros2021probabilistic,guan2021out} are recently introduced to get better results, taking the cropped images as input.
They all need priors to deal with the depth ambiguity.
An extra model such as GMM \cite{bogo2016keep}, GAN \cite{kanazawa2018end,davydov2021adversarial} or VAE \cite{pavlakos2019expressive,kocabas2020vibe} is trained on a large motion capture dataset AMASS \cite{mahmood2019amass} to be an implicit prior.
Other methods search for plausible SMPL parameters that may be close to the ground truth to be an explicit prior \cite{muller2021self,guan2021out}.
We propose a novel annotator based on CLIFF, which takes more than the cropped image as input and calculates the 2D supervision in the full image, and use the SMPL parameter predictions by the pretrained CLIFF annotator as an effective explicit prior, without an extra model and human actors mimicking the predefined poses \cite{muller2021self,kocabas2021spec}.

\section{Approach}\label{sec:approach}

In this section, we first briefly review the commonly-used parametric model SMPL and the baseline method HMR, then propose our model CLIFF, and finally present a novel pseudo-GT annotator for in-the-wild images.

\subsection{SMPL Model}
As a parametric human body model, SMPL \cite{loper2015smpl} provides a differentiable function, which takes dozens of parameters $\Theta = {\left\{ \bm{\theta}, \bm{\beta} \right\}}$ as input, and returns a posed 3D mesh of 6890 vertices $V$, where $\bm{\theta} \in \mathbb{R}^{24\times3}$ and $\bm{\beta} \in \mathbb{R}^{10}$ are the pose and shape parameters, respectively.
The pose parameters $\bm{\theta}$ consist of the global rotation of the root joint (pelvis) with respect to some coordinate system (e.g., the camera coordinate system in our work), and 23 local rotations of other articulated joints relative to their parents along the kinematic tree.
The $k$ joints of interest $J_{3D} \in \mathbb{R}^{k\times3}$ can be obtained by the linear combination of the mesh vertices, $J_{3D} = MV$, where $M$ is a pretrained sparse matrix for these $k$ joints.

\subsection{HMR Model}

\begin{figure}[t]
	\centering
	\includegraphics[width=\textwidth]{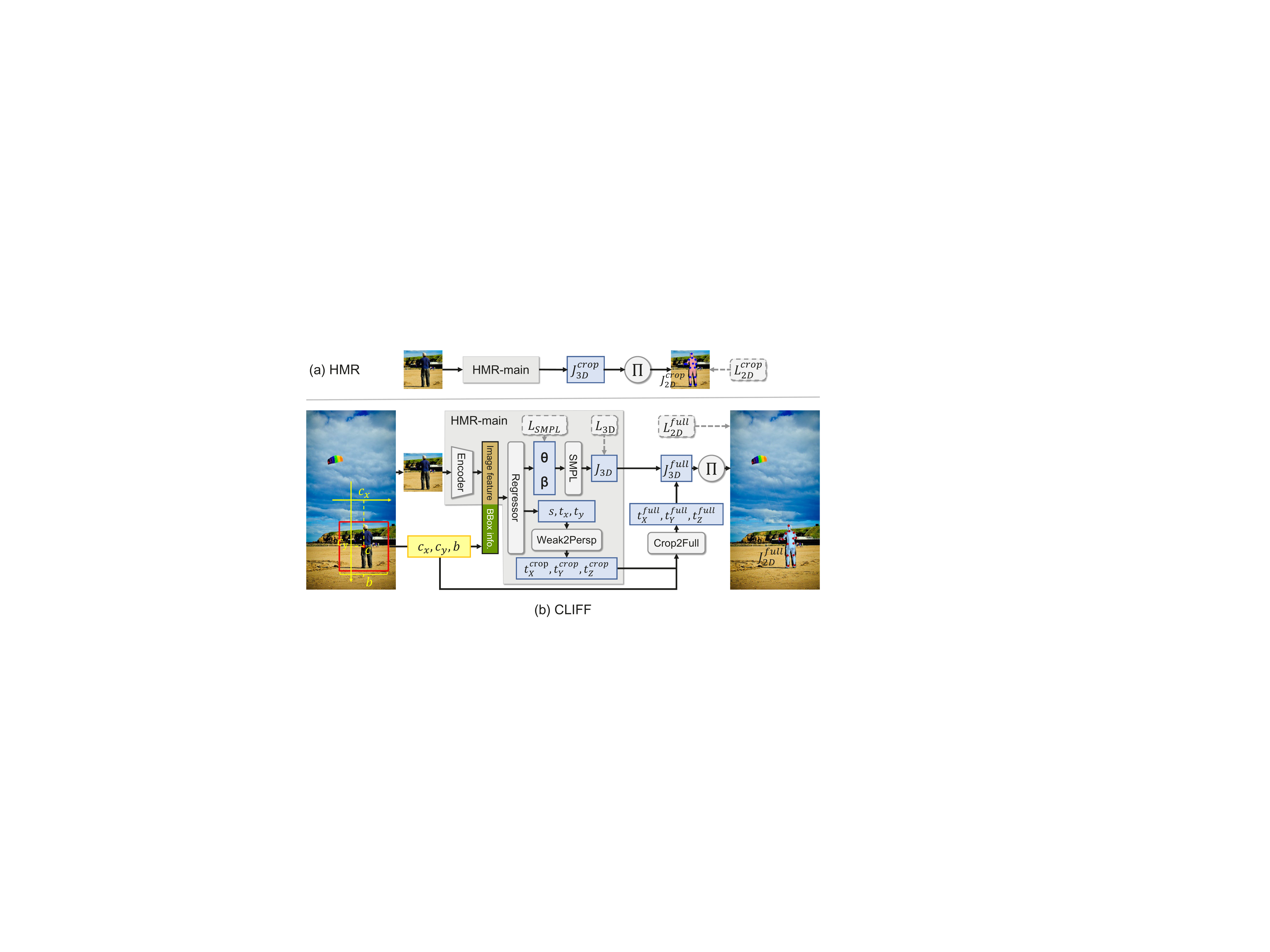}
	\caption {(a) HMR architecture where the detail of its main part ``HMR-main" is shown in the gray box in (b). (b) CLIFF architecture that takes HMR as its backbone and has two modifications: 1) CLIFF takes as input not only the cropped image but also the bounding box information. 2) CLIFF projects the predicted 3D joints onto the original full image plane to compute the 2D reprojection loss, while HMR does this in the cropped image.}
	\label{fig:overview}
\end{figure}


HMR \cite{kanazawa2018end} is a simple and widely-used top-down method for 3D human pose and shape estimation.
Its architecture is shown in Fig. \ref{fig:overview}(a).
A square cropped image is resized to $224 \times 224$ and passed through a convolutional encoder.
Then an iterative MLP regressor predicts the SMPL parameters $\Theta = {\left\{ \bm{\theta}, \bm{\beta} \right\}}$ and weak-perspective projection parameters $P_{weak} = {\left\{ s, t_x, t_y \right\}}$ for a virtual camera $M_{crop}$ with respect to the cropped image (see Fig. \ref{fig:adjust}), where $s$ is the scale parameter, and $t_x$ and $t_y$ are the root translations relative to $M_{crop}$ along the $X$ and $Y$ axes, respectively.
With a predefined large focal length $f_{HMR} = 5000$, $P_{weak}$ can be transformed to perspective projection parameters $P_{persp} = {\left\{ f_{HMR}, \mathbf{t}^{crop} \right\}}$, where $\mathbf{t}^{crop} = [t^{crop}_X, t^{crop}_Y, t^{crop}_Z]$ denotes the root translations relative to $M_{crop}$ along the $X$, $Y$, and $Z$ axes respectively:
\begin{equation}
	\label{eq:convert_crop}
	t^{crop}_X = t_x, \
	t^{crop}_Y = t_y, \
	t^{crop}_Z = \frac{2 \cdot f_{HMR}}{r \cdot s},
\end{equation}
where $r=224$ denotes the side resolution of the resized square crop.

The loss of HMR is defined as:
\begin{equation}
	L^{HMR} = \lambda_{SMPL}L_{SMPL}+\lambda_{3D}L_{3D}+\lambda_{2D}L_{2D}^{crop},
\end{equation}
and its terms are calculated by:
\begin{equation}
	L_{SMPL} = \left \| \Theta - \hat{\Theta} \right \|,
	L_{3D} =   \left \| J_{3D} - \hat{J}_{3D} \right \|,
	L_{2D}^{crop} =   \left \| J_{2D}^{crop} - \hat{J}_{2D}^{crop} \right \|,
\end{equation}
where $\hat{\Theta}$, $\hat{J}_{3D}$, and $\hat{J}_{2D}^{crop}$ are the ground truth, and the predicted 2D keypoints in the cropped image are obtained by the perspective projection $\mathrm{\Pi}$:
\begin{equation}
	J_{2D}^{crop} = \mathrm{\Pi} J_{3D}^{crop}
				  = \mathrm{\Pi} (J_{3D} + \mathbf{1} \mathbf{t}^{crop}),
\end{equation}
where $\mathbf{1} \in \mathbb{R}^{k\times1}$ is an all-ones matrix.

\subsection{CLIFF Model}

\begin{figure}[t]
	\centering
	\includegraphics[width=\textwidth]{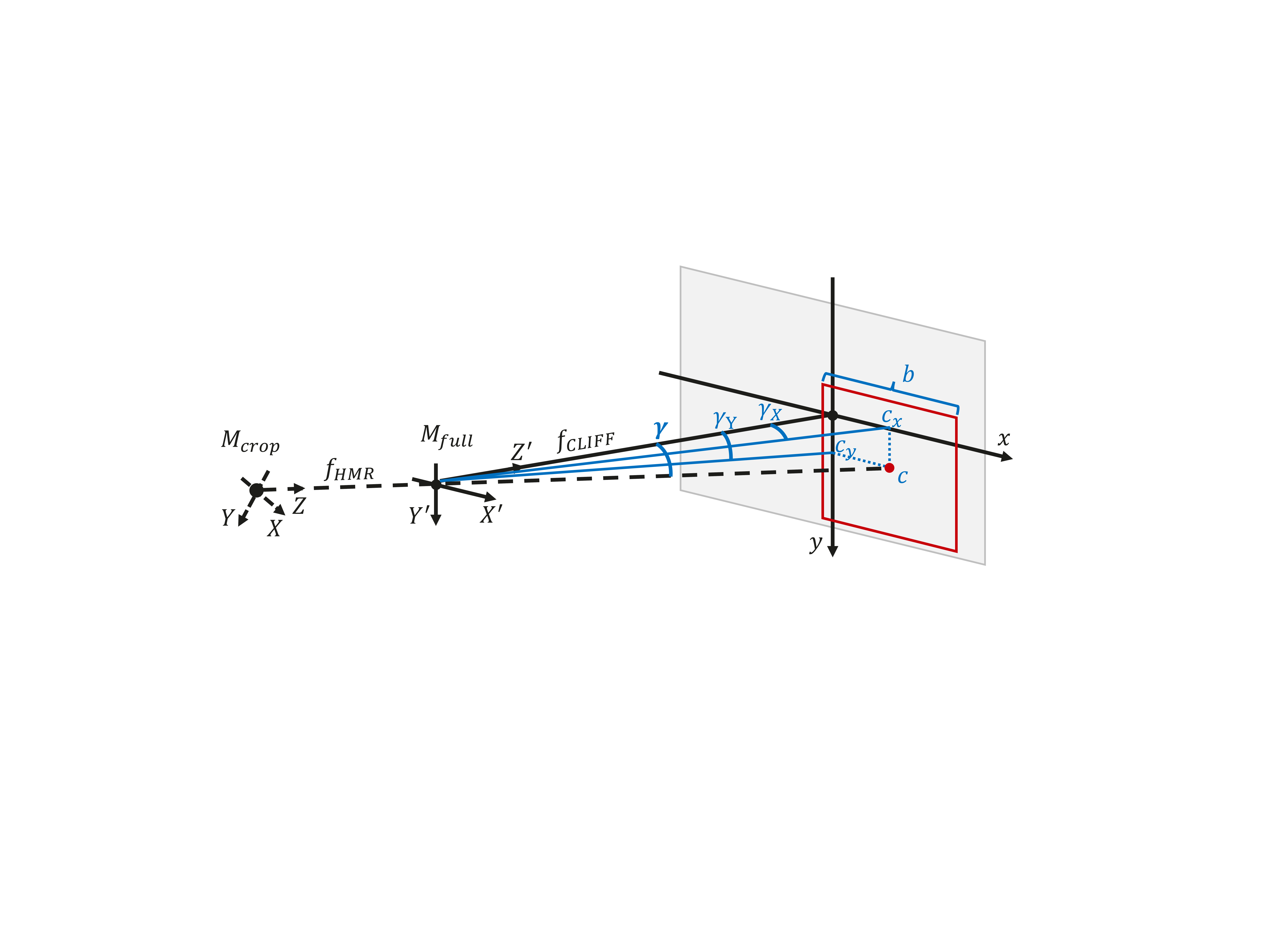}
	\caption {The geometric relation between the virtual camera $M_{crop}$ for the cropped image (the red rectangle) and the original camera $M_{full}$ for the full image.}
	\label{fig:adjust}
\end{figure}

As described above, previous top-down methods take only the cropped image as input, and calculate the reprojection loss in the cropped image, which may lead to inaccurate predictions, as emphasized in Section \ref{sec:intro}.
To address this problem, we take HMR as the baseline and propose to make two modifications to build CLIFF, as shown in Fig. \ref{fig:overview}(b).

First, CLIFF takes more holistic features as input.
Besides the encoded image feature, the additional bounding box information $I_{bbox}$ of the cropped region is also fed to the regressor,
\begin{equation}
I_{bbox} = [\frac{c_x}{f_{CLIFF}}, \frac{c_y}{f_{CLIFF}}, \frac{b}{f_{CLIFF}}],
\end{equation}
where $(c_x, c_y)$ is its location relative to the full image center, $b$ its original size, and $f_{CLIFF}$ the focal length of the original camera $M_{full}$ used in CLIFF (see Fig. \ref{fig:adjust}).
Besides the effect of normalization, taking $f_{CLIFF}$ as the denominator gives geometric meanings to the first two terms in $I_{bbox}$:
\begin{equation}\begin{aligned}
\tan \gamma_X & = \frac{c_x}{f_{CLIFF}}, \\
\tan \gamma_Y & = \frac{c_y}{f_{CLIFF}},
\end{aligned}\end{equation}
where $\bm{\gamma} = [\gamma_X$, $\gamma_Y, 0]$ is the transformation angle from $M_{crop}$ to the original camera $M_{full}$ coordinate system with respect to the full image, as shown in Fig. \ref{fig:adjust}.
Therefore, fed with $I_{bbox}$ as part of the input, the regressor can make the transformation implicitly to predict the global rotation relative to $M_{full}$, which also bring benefits to the articulated pose estimation.
As for the focal length, we use the ground truth if it is known; otherwise we approximately estimate its value as $f_{CLIFF} = \sqrt{w^2 + h^2}$, where $w$ and $h$ are the width and height of the full image respectively, corresponding to a diagonal Field-of-View of 55° for $M_{full}$, following the previous work \cite{kissos2020beyond}.

Second, CLIFF calculates the reprojection loss in the full image instead of the cropped one.
The root translation is transformed from $M_{crop}$ to $M_{full}$:
\begin{equation}\begin{aligned}
	\label{eq:convert_full}
	t_X^{full} & = t_X^{crop} + \frac{2 \cdot c_x}{b \cdot s}, \\
	t_Y^{full} & = t_Y^{crop} + \frac{2 \cdot c_y}{b \cdot s}, \\
	t_Z^{full} & = t_Z^{crop} \cdot \frac{f_{CLIFF}}{f_{HMR}} \cdot \frac{r}{b},
\end{aligned}\end{equation}
where $\mathbf{t}^{full}=[t^{full}_X, t^{full}_Y, t^{full}_Z]$ denotes the root translations relative to $M_{full}$ along the $X^{'}$, $Y^{'}$, and $Z^{'}$ axes, respectively. The derivation can be found in our supplementary materials.
Then we project the predicted 3D joints onto the full image plane:
\begin{equation}
	J_{2D}^{full} = \mathrm{\Pi} J_{3D}^{full}
			      = \mathrm{\Pi} (J_{3D} + \mathbf{1} \mathbf{t}^{full}),
\end{equation}
to compute the 2D reprojection loss:
\begin{equation}
	L_{2D}^{full} = \left \| J_{2D}^{full} - \hat{J}_{2D}^{full} \right \|,
\end{equation}
where the ground truth $\hat{J}_{2D}^{full}$ is also relative to the full image center.
Finally, the total loss of CLIFF is calculated by:
\begin{equation}\begin{aligned}
	L^{CLIFF} = \lambda_{SMPL}L_{SMPL}+\lambda_{3D}L_{3D}+\lambda_{2D}L_{2D}^{full}.
\end{aligned}\end{equation}

The predicted 2D keypoints $J_{2D}^{full}$ share a similar projection process and perspective distortion with the person in the image, especially when the focal length $f_{CLIFF}$ is close to its ground truth. Thus, the corresponding loss $L_{2D}^{full}$ can correctly supervise CLIFF to make more accurate prediction of 3D human poses, especially the global rotation, which is demonstrated in our experiments.

\subsection{CLIFF Annotator}\label{sec:anno}

\begin{figure}[t]
	\centering
	\includegraphics[width=\textwidth]{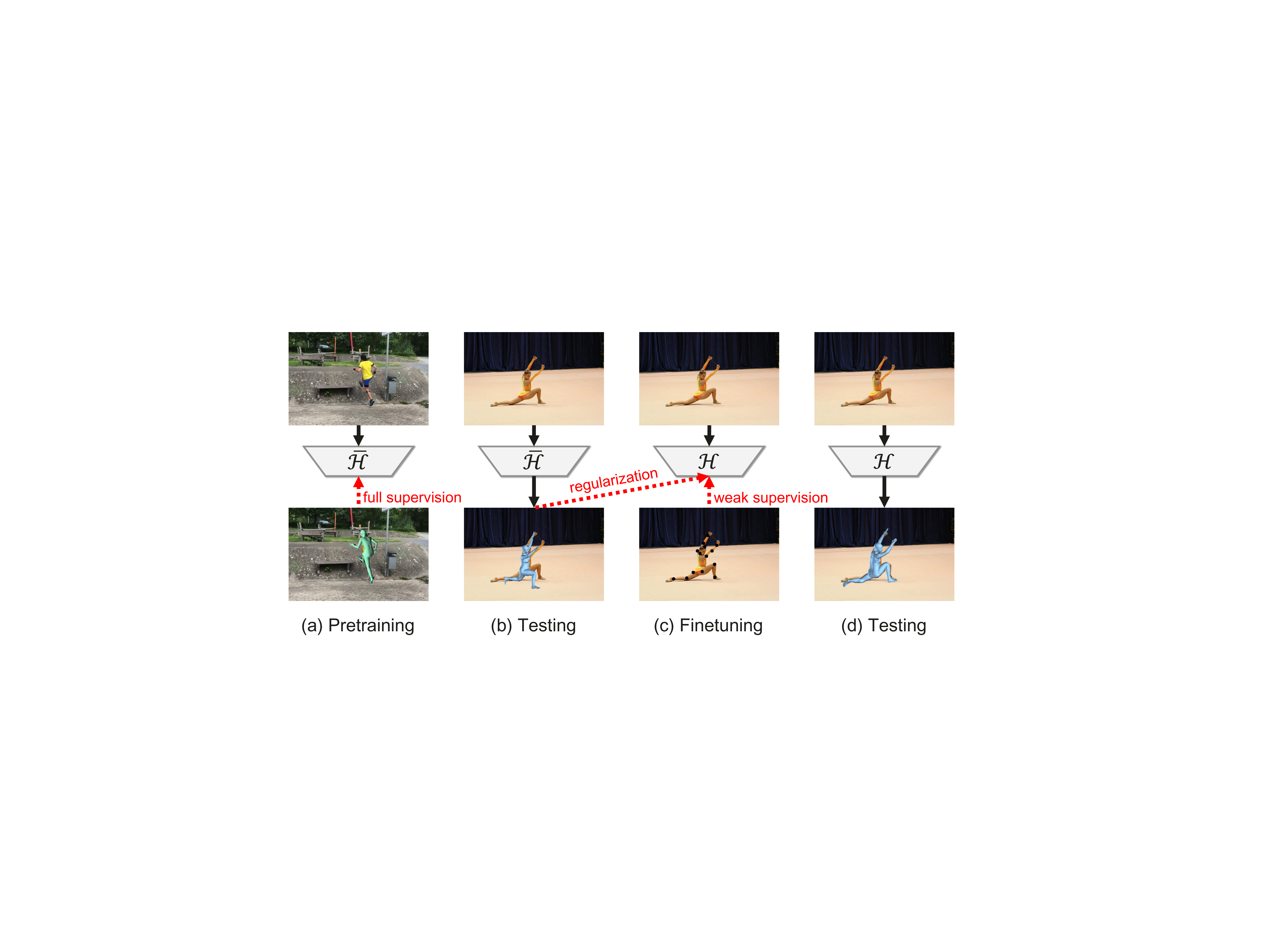}
	\caption {\textbf{Our pipeline to generate 3D annotations for a 2D dataset.} (a) Pretrain CLIFF model $\bar{\mathcal{H}}$ with full supervision on available datasets. (b) Then test $\bar{\mathcal{H}}$ on the target dataset to predict SMPL parameters $\bar{\Theta}$. (c) Finetune $\bar{\mathcal{H}}$ with 2D keypoints as weak supervision and $\bar{\Theta}$ as regularization, obtaining the updated model $\mathcal{H}$. (d) Finally, test $\mathcal{H}$ on the target dataset to infer SMPL parameters as pseudo-GT.}
	\label{fig:annotation}
\end{figure}

Full supervision from 3D ground truth (particularly the SMPL parameters) is crucial for regression-based methods to improve their performances on 3D human pose and shape estimation \cite{kolotouros2019learning}. However, these annotations are scarce for in-the-wild datasets, since they require specialized devices and cost a lot of time and labor \cite{ionescu2013human3,von2018recovering}. Recently, CNN-based pseudo-GT annotators \cite{kolotouros2019learning,joo2021exemplar,moon2020neuralannot} are proposed to address this problem. However, their base models are agnostic to the person locations in full frames, and thus they produce inaccurate annotations, especially for the global rotations, as mentioned in Section \ref{sec:intro}.

Hence, we propose an annotator based on CLIFF, which is fed and supervised with global-location-aware information, and thus produces better global rotation and articulated pose annotations simultaneously.
As shown in Fig. \ref{fig:annotation}, there are four steps in our pipeline to annotate an in-the-wild dataset with only 2D keypoint ground truth.
\begin{enumerate}
\item We pretrain the CLIFF annotator $\bar{\mathcal{H}}$ on several datasets with ground-truth SMPL parameters, including 3D datasets and 2D datasets with pseudo-GT generated by EFT \cite{joo2021exemplar}.
The pretrained weights serve as an implicit prior from these various datasets for the following optimization in Step 3.

\item Test the pretrained model $\bar{\mathcal{H}}$ on the target dataset to predict SMPL parameters $\bar{\Theta}$.
Although these predictions may not be accurate, they can be an explicit prior to guide the optimization, and cost little without the need for crowd-sourced participants to mimick some predefined poses \cite{muller2021self,kocabas2021spec}.

\item Finetune the pretrained model $\bar{\mathcal{H}}$ on the target dataset, using ground-truth 2D keypoints as weak supervision and $\bar{\Theta}$ as regularization, to get the updated annotator $\mathcal{H}$.
Due to the depth ambiguity, these 2D keypoints are insufficient to supervise the optimization to recover their 3D ground truth. Thus, the priors are very important because they can prevent $\mathcal{H}$ from overfitting to these 2D keypoints and from offering implausible solutions.

\item Test $\mathcal{H}$ on the target dataset to get SMPL parameter predictions $\Theta$ as the final pseudo-GT. In our experiment, the reconstructed 3D meshes from these pseudo-GT are pixel-aligned to their 2D evidences, and also perceptually realistic which can be confirmed from novel views.
\end{enumerate}

Compared with other annotators whose priors come from extra models \cite{bogo2016keep,pavlakos2019expressive} trained on another large motion capture dataset AMASS \cite{mahmood2019amass}, the CLIFF annotator contains strong priors that are efficient to obtain with no need for an extra model and AMASS.
More importantly, based on CLIFF, the annotator produces much better pseudo-GT which is very helpful to boost the training performance as shown in the experiments.

\subsection{Implementation details}
We train CLIFF for 244K steps with batch size 256, using the Adam optimizer \cite{kingma2015adam}.
The learning rate is set to $1 \times e^{-4}$ and reduced by a factor of 10 in the middle.
The image encoder is pretrained on ImageNet \cite{deng2009imagenet}.
The cropped images are resized to $224 \times 224$, preserving the aspect ratio.
Data augmentation includes random rotations and scaling, horizontal flipping, synthetic occlusion \cite{sarandi2018robust}, and random cropping \cite{joo2021exemplar,rockwell2020full}.
To annotate an in-the-wild dataset, we train the CLIFF annotator for 30 epochs with learning rate $5 \times e^{-5}$ but no data augmentation.
We use MindSpore \cite{ms} and PyTorch \cite{paszke2019pytorch} for the implementation.

\section{Experiments and Results}

\subsubsection{Datasets.}
Following previous work, we train CLIFF on a mixture of 3D datasets (Human3.6M \cite{ionescu2013human3} and MPI-INF-3DHP \cite{mehta2017monocular}), and 2D datasets (COCO \cite{lin2014microsoft} and MPII \cite{andriluka20142d}) with pseudo-GT provided by the CLIFF annotator.
The evaluation is performed on three datasets: 1) 3DPW \cite{von2018recovering}, an in-the-wild dataset with 3D annotations from IMU devices; 2) Human3.6M, an indoor dataset providing 3D ground truth from optical markers with a multi-view setup; 3) AGORA \cite{patel2021agora}, a synthetic dataset with highly accurate 3D ground truth.
We use the 3DPW and AGORA training data when conducting experiments on them respectively.

\subsubsection{Evaluation metrics.}
The three standard metrics in our experiments are briefly described below.
They all measure the Euclidean distances (in millimeter (mm)) of 3D points between the predictions and ground truth.
\begin{itemize}
	\item [] \textbf{MPJPE}
	(Mean Per Joint Position Error) first aligns the predicted and ground-truth 3D joints at the pelvis, and then calculates their distances, which comprehensively evaluates the predicted poses and shapes, including the global rotations.
	\item [] \textbf{PA-MPJPE}
	(Procrustes-Aligned Mean Per Joint Position Error, or reconstruction error) performs Procrustes alignment before computing MPJPE, which mainly measures the articulated poses, eliminating the discrepancies in scale and global rotation.
	\item [] \textbf{PVE}
	(Per Vertex Error, or MVE used in the AGORA evaluation) does the same alignment as MPJPE at first, but calculates the distances of vertices on the human mesh surfaces.
\end{itemize}

\begin{table}[t]\scriptsize
	\centering
	\caption{Performance comparison between CLIFF and state-of-the-art methods on 3DPW, Human3.6M and AGORA}
	\label{table:sota}
	\begin{tabular}{clccccccc}
		\toprule
		\multicolumn{1}{l}{} &      & \multicolumn{3}{c}{3DPW}                                      & \multicolumn{2}{c}{Human3.6M}                & \multicolumn{2}{c}{AGORA}                   \\
		\cmidrule(lr){3-5}\cmidrule(lr){6-7}\cmidrule(lr){8-9}
		\multicolumn{1}{l}{} & Method          &
		\multicolumn{1}{c}{MPJPE $\downarrow$} & \multicolumn{1}{c}{PA-MPJPE $\downarrow$} & \multicolumn{1}{c}{PVE $\downarrow$} & \multicolumn{1}{c}{MPJPE $\downarrow$} & \multicolumn{1}{c}{PA-MPJPE $\downarrow$} & \multicolumn{1}{c}{MPJPE $\downarrow$} & \multicolumn{1}{c}{MVE $\downarrow$}  \\
		\midrule
		\multirow{4}{*}{\rotatebox{90}{video}}
		& HMMR \cite{kanazawa2019learning}	& 116.5 & 72.6   & -    & -    & 56.9   & -    & -     \\
		& TCMR \cite{choi2021beyond}		& 86.5 & 52.7   & 102.9 & -    & -      & -    & -     \\
		& VIBE \cite{kocabas2020vibe} 		& 82.7 & 51.9   & 99.1 & 65.6 & 41.1   & -    & -     \\
		& MAED \cite{wan2021encoder} 		& 79.1 & 45.7   & 92.6 & 56.4 & 38.7   & -    & -     \\
		\midrule
		\multirow{5}{*}{\rotatebox{90}{model-free}}
		& I2L-MeshNet \cite{moon2020i2l}     & 93.2 & 58.6   & 110.1 & -    & -      & -    & -     \\
		& Pose2Mesh \cite{choi2020pose2mesh}       & 89.5 & 56.3   & 105.3 & 64.9 & 46.3   & -    & -     \\
		& HybrIK \cite{li2021hybrik}         & 80.0 & 48.8   & 94.5 & 54.4 & 34.5   & -    & -     \\
		& METRO \cite{lin2021end}           & 77.1 & 47.9   & 88.2 & 54.0 & 36.7   & -    & -     \\
		& Graphormer \cite{lin2021mesh} & 74.7 & 45.6   & 87.7 & 51.2 & 34.5   & -    & -     \\
		\midrule
		\multirow{9}{*}{\rotatebox{90}{model-based}}
		& HMR \cite{kanazawa2018end}  & 130.0 & 81.3   & -    & -    & 56.8   & 180.5 & 173.6 \\
		& SPIN \cite{kolotouros2019learning} & 96.9 & 59.2   & 116.4 & -    & 41.1   & 153.4 & 148.9 \\
		& SPEC \cite{kocabas2021spec} & 96.5 & 53.2   & 118.5 & -    & -      & 112.3 & 106.5 \\
		& HMR-EFT \cite{joo2021exemplar}  & 85.1 & 52.2   & 98.7 & 63.2 & 43.8   & 165.4 & 159.0 \\
		& PARE \cite{kocabas2021pare} & 79.1 & 46.4   & 94.2 & -    & -      & 146.2 & 140.9 \\
		& ROMP \cite{sun2021monocular} & 76.7 & 47.3   & 93.4 & -    & -      & 116.6 & 113.8 \\
		\cmidrule{2-9}
		& CLIFF (Res-50) & 72.0 & 45.7   & 85.3 & 50.5 & 35.1 & 91.7 & 86.3  \\
		& CLIFF (HR-W48) & \B 69.0 & \B 43.0 & \B 81.2 & \B 47.1 & \B 32.7 & \B 81.0 & \B 76.0  \\
		\bottomrule
	\end{tabular}
\end{table}

\begin{figure}[t]
	\centering
	\includegraphics[width=\textwidth]{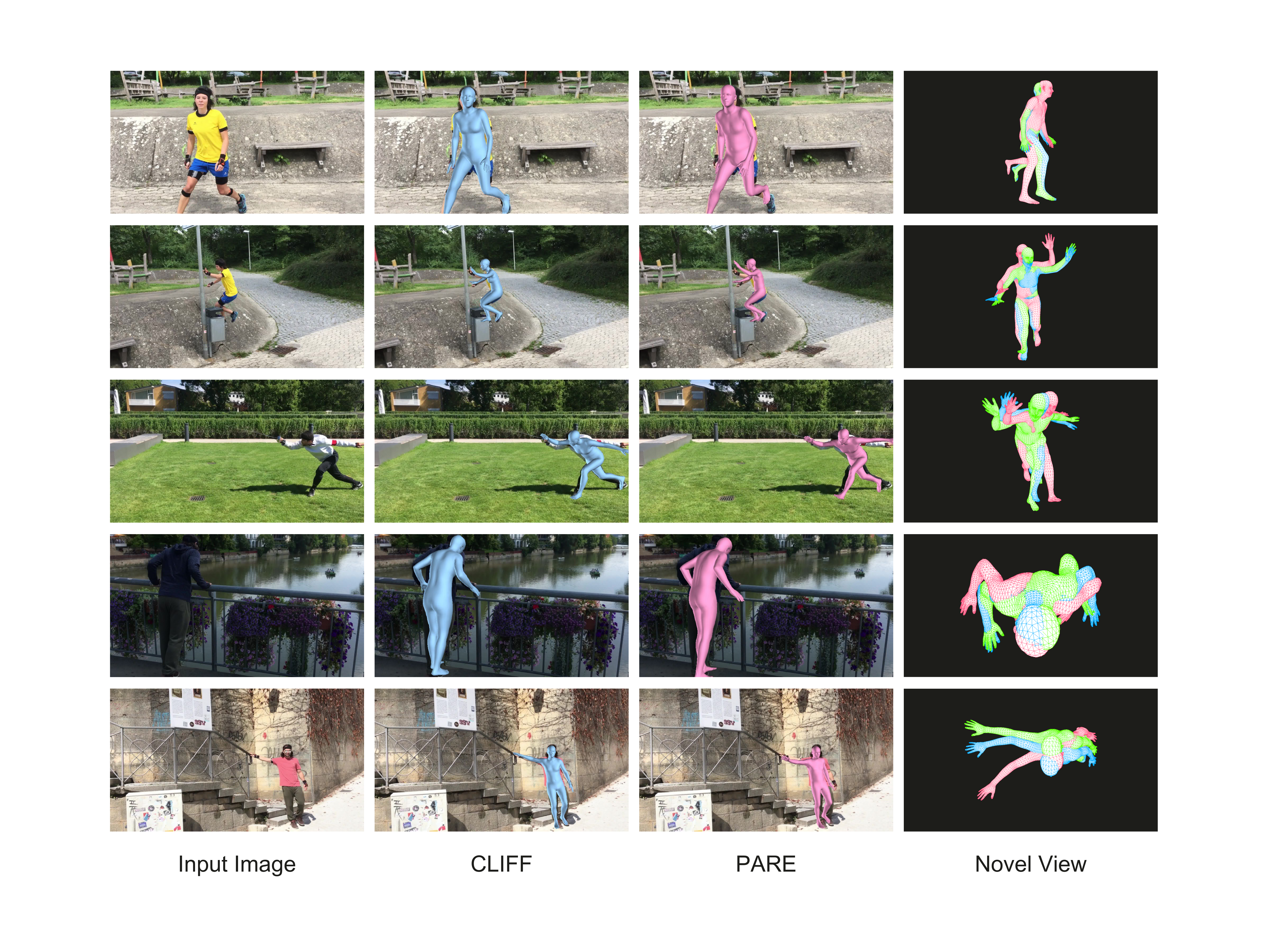}
	\caption {\textbf{Qualitative comparison on 3DPW.} From left to right: input images, CLIFF predictions, PARE predictions, and their visualizations from novel views (green for the ground truth, blue for CLIFF, and red for PARE).}
	\label{fig:cliff_comp}
\end{figure}

\begin{table}[t]\scriptsize
	\begin{minipage}[t]{0.45\linewidth}\centering
		\caption{ Ablation study of CLIFF on Human3.6M }
		\label{table:ablation}
		\begin{tabular}{lcc}
			\toprule
			Method & MPJPE $\downarrow$ & PA-MPJPE $\downarrow$ \\
			\midrule
			CLIFF w/o CI \& CS 			& 85.2 & 54.5 \\
			CLIFF w/o CI			 	& 84.0 & 52.4 \\
			CLIFF			& \B 81.4 & \B 52.1 \\
			\bottomrule
		\end{tabular}
	\end{minipage}
	\hfill
	\begin{minipage}[t]{0.49\linewidth}\centering
		\caption{ Direct comparison among pseudo-GT annotators on 3DPW }
		\label{table:anno-direct}
		\begin{tabular}{lccc}
			\toprule
			Annotator & MPJPE $\downarrow$ & PA-MPJPE $\downarrow$  & PVE $\downarrow$ \\
			\midrule
			ProHMR \cite{kolotouros2021probabilistic} 				&    - & 52.4 & - \\
			BOA \cite{guan2021bilevel}				& 77.2 & 49.5 & - \\
			EFT \cite{joo2021exemplar} 				&    - & 49.3 & - \\
			DynaBOA \cite{guan2021out}				& 65.5 & 40.4 & 82.0 \\
			Pose2Mesh \cite{choi2020pose2mesh}			& 65.1 & 34.6 & - \\
			\midrule
			CLIFF (Ours)		& \B 52.8 & \B 32.8 & \B 61.5 \\
			\bottomrule
		\end{tabular}
	\end{minipage}
\end{table}

\subsection{Comparisons with State-of-the-art Methods}

\subsubsection{Quantitative results.}
We compare CLIFF with prior arts, including video-based methods \cite{kanazawa2019learning,choi2021beyond,kocabas2020vibe,wan2021encoder} that exploit temporal information, and frame-based ones \cite{choi2020pose2mesh,li2021hybrik,lin2021mesh,kolotouros2019learning} that process each frame independently.
They could be model-based \cite{kanazawa2018end,kocabas2021pare} or model-free \cite{moon2020i2l,lin2021end}, and most of them are top-down methods, except for one bottom-up \cite{sun2021monocular}.
As shown in Table \ref{table:sota}, CLIFF outperforms them by significant margins in all metrics on these three evaluation datasets.
With the same image encoder backbone (ResNet-50 \cite{he2016identity}) and similar computation cost, CLIFF beats its baseline HMR-EFT, reducing the errors by more than 13mm on MPJPE and PVE.
In the case of similar PA-MPJPE to other methods, CLIFF still has lower MPJPE and PVE, since it has a better global rotation estimation.
With HRNet-W48 \cite{sun2019deep}, CLIFF has better performance and distinct advantages over previous state-of-the-art, including METRO \cite{lin2021end} and Mesh Graphormer \cite{lin2021mesh} which have similar image encoder backbones (HRNet-W64) and transformer-based architectures \cite{vaswani2017attention}.
CLIFF reaches \textit{the first place} on \href{https://agora-evaluation.is.tuebingen.mpg.de}{the AGORA leaderboard} (the SMPL-Algorithms track) way ahead of other methods (whose results are from the leaderboard).


\subsubsection{Qualitative results.}
As shown in Fig. \ref{fig:cliff_comp}, we compare CLIFF with PARE \cite{kocabas2021pare} on the 3DPW testing data, which is one of the best cropped-image-based methods.
We render the reconstructed meshes using the original camera with ground-truth intrinsic parameters.
Even though accurate articulated poses can also be obtained by PARE, we can see clear pixel-misalignment of its results overlaid on the images, suffering from its inferior global rotation estimation.
From the novel viewpoints, we can see that the predicted meshes by CLIFF overlay with the ground truth better than those by PARE, thanks to its more accurate global rotation estimation.

\begin{figure}[t]
	\centering
	\includegraphics[width=\textwidth]{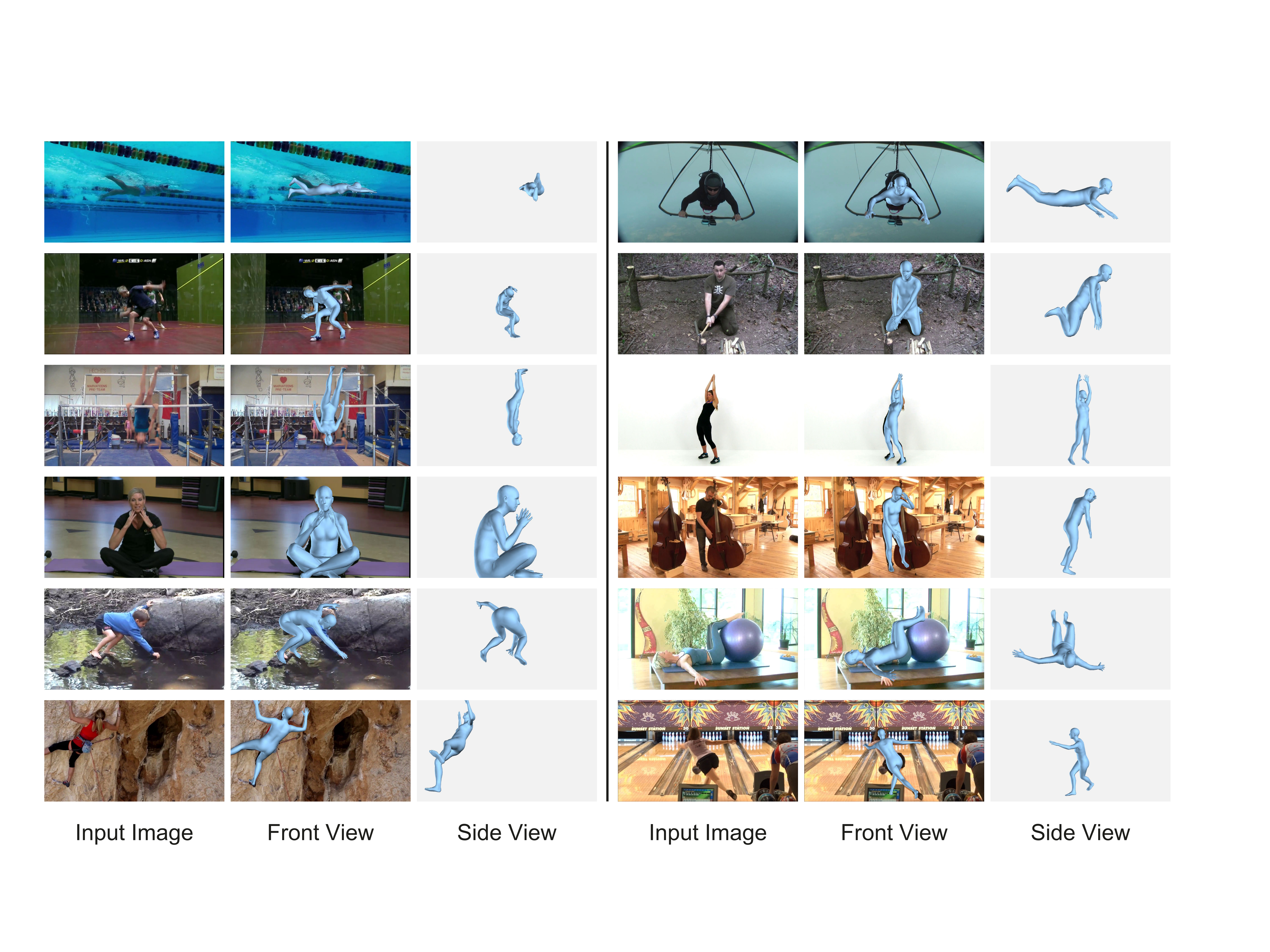}
	\caption {\textbf{Qualitative results of the CLIFF annotator on the MPII dataset.} From left to right: input images, the front view and side view of the pseudo-GT.}
	\label{fig:anno_results}
\end{figure}

\begin{table}[t]\scriptsize
	\centering
	\caption{Indirect comparison among pseudo-GT annotators by training on COCO and then evaluating on 3DPW and Human3.6M}
	\label{table:anno-indirect}
	\begin{tabular}{lccccc}
		\toprule
		& \multicolumn{3}{c}{3DPW}
		& \multicolumn{2}{c}{Human3.6M}                        \\
		\cmidrule(lr){2-4}\cmidrule(lr){5-6}
		Annotator   &
		\multicolumn{1}{c}{MPJPE $\downarrow$} & \multicolumn{1}{c}{PVE $\downarrow$} & \multicolumn{1}{c}{PA-MPJPE $\downarrow$} & \multicolumn{1}{c}{MPJPE $\downarrow$} & \multicolumn{1}{c}{PA-MPJPE $\downarrow$}  \\
		\midrule
		SPIN \cite{kolotouros2019learning}          & 101.2 & 119.2 & 65.2 & 115.0 & 66.9 \\
		EFT \cite{joo2021exemplar}           &  98.8 & 115.8 & 62.0 & 110.9 & 60.7 \\
		\midrule
		CLIFF (Ours)  & \B 85.4 & \B 100.5 & \B 53.6 & \B 96.1 & \B 54.8 \\
		\bottomrule
	\end{tabular}
\end{table}

\subsection{Ablation Study}
We take HMR as the baseline and make two modifications to build CLIFF: additional input of the bounding box information (CI, denoting the CLIFF Input), and the 2D reprojection loss calculated in the full frame for supervision (CS, denoting the CLIFF Supervision).
As shown in Table \ref{table:ablation}, we conduct an ablation study on Human3.6M, since it has accurate 3D ground truth.
Without CI providing enough information, MPJPE increases significantly, indicating a worse global rotation estimation.
It causes larger errors when we also drop CS that can guide CLIFF to better predictions.
This study validates our intuition that global-location-aware information helps the model predict the global rotation and obtain more accurate articulated poses.

\subsection{Annotator Comparisons}

\subsubsection{Direct comparison.}
In Table \ref{table:anno-direct}, we directly compare different pseudo-GT annotators on 3DPW, for it is an in-the-wild dataset with ground-truth SMPL parameters.
The CLIFF annotator outperforms other methods in all metrics.
Even with similar PA-MPJPE to Pose2Mesh \cite{choi2020pose2mesh}, CLIFF reduces MPJPE by 12.3mm, and PVE by 20.5mm.
Compared to EFT \cite{joo2021exemplar} that finetunes a pretrained model on each example, the CLIFF annotator is trained in a mini-batch manner, which helps it maintain the implicit prior all the way.
With the additional explicit prior, there is no need for our annotator to choose a generic stopping criterion carefully.
It only takes about 30 minutes for the CLIFF annotator to annotate the whole 35,515 images with 4 Tesla V100 GPUs (the finetuning and final testing steps described in Section \ref{sec:anno}).

\subsubsection{Indirect comparison.}
We train CLIFF on the COCO dataset with pseudo-GT from different annotators, and show their results on 3DPW and Human3.6M.
The training lasts for 110K steps without learning rate decay.
As shown in Table \ref{table:anno-indirect}, the CLIFF annotator has much better performance than SPIN \cite{kolotouros2019learning} and EFT \cite{joo2021exemplar} (more than 13mm margins on MPJPE and PVE).
It demonstrates that the CLIFF annotator can generate high-quality pseudo-GT for in-the-wild images with only 2D annotations, which helps to improve performances significantly.

\subsubsection{Qualitative results.}
In Fig. \ref{fig:anno_results}, we show pseudo-GT samples generated by the CLIFF annotator.
With good 2D keypoint annotations, the reconstructed meshes are pixel-aligned to the image evidence.
From the side view, we can see that they are also perceptually realistic without obvious artifacts, thanks to the strong priors in the CLIFF annotator.

\section{Discussion}
In Section \ref{sec:approach}, we show how to build CLIFF based on HMR by making two modifications.
We believe that the idea can also be applied to many other methods.
First, it can benefit regression-based top-down methods that work on the cropped-region features (e.g., image, keypoint, edge, and silhouette).
As for bottom-up methods that treat all the subjects without distinction of their different locations, we can take another form to encode the location information, for example, a location map which consists of a normalized coordinate for each pixel.
Going beyond 3D human pose estimation, we can apply the idea to other 3D tasks that involve object global rotations (e.g., 3D object detection and 6-DoF object pose estimation).
Even when there are perfect 3D annotations and thus no need for the 2D reprojection loss calculated in the full image, it is still important to take the global-location-aware information as input.

\section{Conclusion}

Although translation invariance is a key factor for CNN models to succeed in computer vision, we argue that the global location information in full frames matters in 3D human pose and shape estimation, and the global rotations cannot be accurately inferred when only using cropped images.
To address this problem, we propose CLIFF by feeding and supervising the model with global-location-aware information.
CLIFF takes not only the cropped image but also its bounding box information as input.
It calculates the 2D reprojection loss in the full image instead of the cropped one, projecting the predicted 3D joints in a way similar to that of the person projected in the image.
Moreover, based on CLIFF, we propose a novel pseudo-GT annotator for in-the-wild 2D datasets, which generates high-quality 3D annotations to help regression-based models boost their performances.
Extensive experiments on popular benchmarks show that CLIFF outperforms state-of-the-art methods by a significant margin and reaches the first place on the AGORA leaderboard (the SMPL-Algorithms track).

\clearpage

%
%
\bibliographystyle{splncs04}
\bibliography{egbib, egbib_supp}



\title{Supplementary Materials \\
	\scriptsize ~ \\
	\normalsize CLIFF: Carrying Location Information in Full Frames \\
	into Human Pose and Shape Estimation}

\titlerunning{CLIFF}

\author{
Zhihao Li \and
Jianzhuang Liu \and
Zhensong Zhang \and
Songcen Xu \and
Youliang Yan
}

\authorrunning{Z. Li et al.}

\institute{
Huawei Noah's Ark Lab
\email{\{zhihao.li,liu.jianzhuang,zhangzhensong,xusongcen,yanyouliang\}@huawei.com}}
\maketitle

In these supplementary materials, we first derive Equation 7 given in the main paper, then take a further discussion about the CLIFF input and its performance when applied to videos, and finally provide more details about the CLIFF annotator.

\section{Derivation of Equation 7}

\begin{figure}[t]
	\centering
	\includegraphics[width=0.8\textwidth]{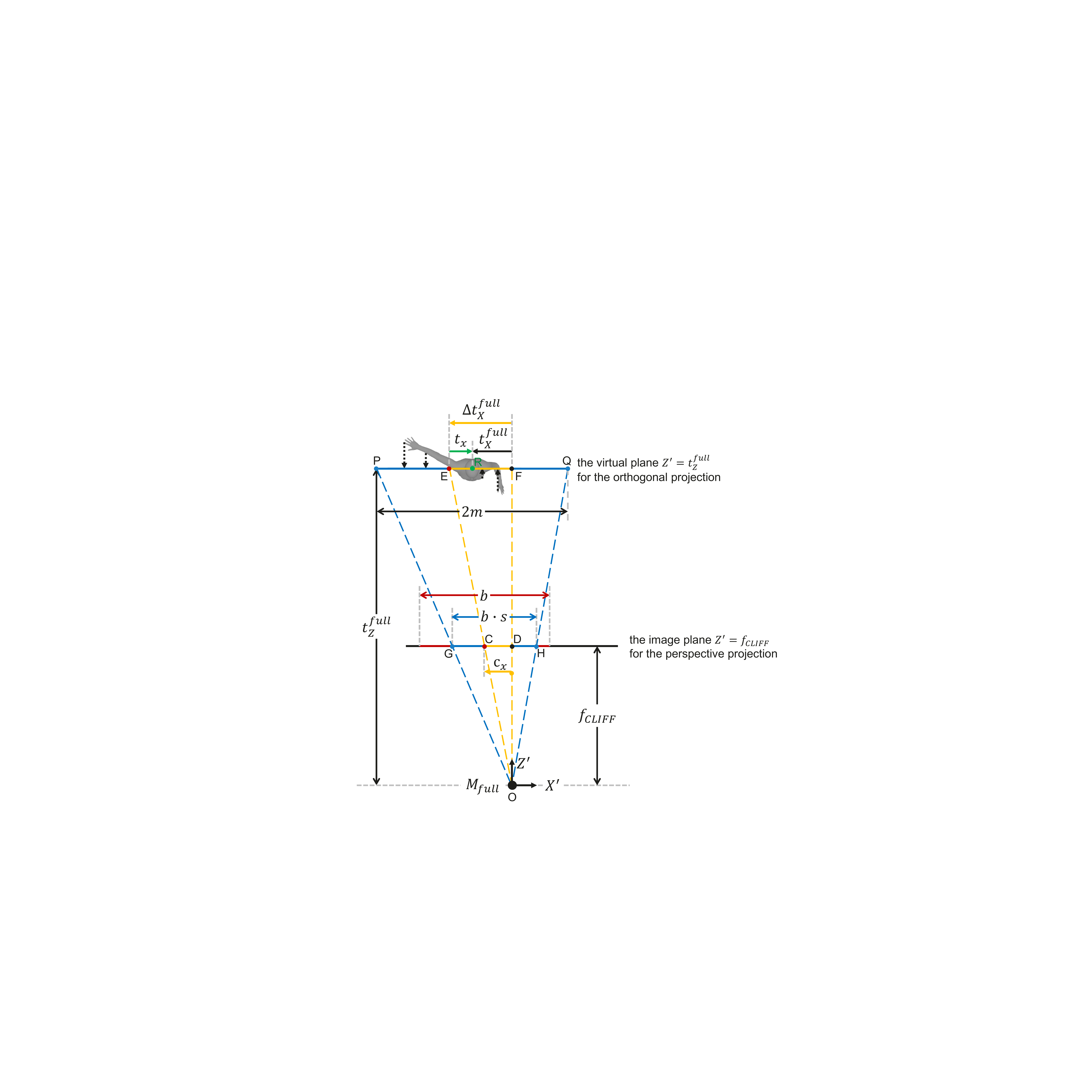}
	\caption {The transformation from weak-perspective projection to perspective projection (bird's eye view). A weak-perspective projection can be regarded as an orthogonal projection followed by a perspective projection. Best viewed in color.}
	\label{fig:derivation}
\end{figure}

\subsubsection{Equation 7.}
CLIFF computes the reprojection loss in the full frame instead of the cropped image, so we need to calculate the root translation $\mathbf{t}^{full}=[t^{full}_X, t^{full}_Y, t^{full}_Z]$ in the coordinate system of the original camera $M_{full}$.
Inserting Equation 1 into Equation 7, we have:
\begin{equation}\tag{7.1}\begin{aligned}
	\label{eq:convert_full}
	t_X^{full} & = t_x + \frac{2 \cdot c_x}{b \cdot s}, \\
	t_Y^{full} & = t_y + \frac{2 \cdot c_y}{b \cdot s}, \\
	t_Z^{full} & = \frac{2 \cdot f_{CLIFF}}{b \cdot s},
\end{aligned}\end{equation}
where $s$, $t_x$, and $t_y$ are the scale and translation parameters of the weak-perspective projection, $(c_x, c_y)$ is the crop location relative to the full image center, $b$ is the size of the original crop (detection result), and $f_{CLIFF}$ is the focal length of the original camera.
See Fig. \ref{fig:derivation} for the illustration.

A weak-perspective projection can be regarded as an orthogonal projection followed by a perspective projection \cite{shimshoni1999geometric}.
As shown in Fig. \ref{fig:derivation}, the human body is first projected (parallel to the $Z^{'}$ axis) onto the virtual plane $Z^{'}=t_Z^{full}$, and then onto the image plane $Z^{'}=f_{CLIFF}$ by a perspective projection.
A T-pose human body of the mean shape is about $1.8m \times 1.8m$ ($m$ denoting meters).
We enclose it with a slightly enlarged box $B$ of size $2m \times 2m$, and align the center of $B$ at the root of the human body (the green point R in Fig. \ref{fig:derivation}).
$B$ is projected to be a square region of size $b \cdot s$ in the image.
Since the two triangles $\triangle$OGH and $\triangle$OPQ in Fig. \ref{fig:derivation} (in blue) are similar, we have:
\begin{equation}\tag{7.2}\label{7.2}
	\frac{2}{t_Z^{full}} = \frac{b \cdot s}{f_{CLIFF}}, ~~~
	t_Z^{full} = \frac{2 \cdot f_{CLIFF}}{b \cdot s}.
\end{equation}
Note that here $b$ and $f_{CLIFF}$ are in pixels, and $t_Z^{full}$ is in meters.

Let D (the projection of F) be the image center, and C (the projection of E) be the crop (i.e., detection result) center.
Then the root translation of the human body along the $X^{'}$ axis is calculated by:
\begin{equation}\tag{7.3}
	t_X^{full} = t_x + \Delta t_X^{full},
\end{equation}
where $\Delta t_X^{full}$ is the $X^{'}$ coordinate of point E.
Since the two triangles $\triangle$OCD and $\triangle$OEF in Fig. \ref{fig:derivation} (in yellow) are similar, we have:
\begin{equation}\tag{7.4}\label{7.4}
	\frac{\Delta t_X^{full}}{t_Z^{full}} = \frac{c_x}{f_{CLIFF}}.
\end{equation}
Combining Equations \ref{7.2} and \ref{7.4}, we obtain:
\begin{equation}\tag{7.5}
	\Delta t_X^{full} = \frac{2 \cdot c_x}{b \cdot s}.
\end{equation}
Similarly, it also holds for the root translation along the $Y^{'}$ axis:
\begin{equation}\tag{7.6}\begin{aligned}
	t_Y^{full} & = t_y + \Delta t_Y^{full} \\
			   & = t_y + \frac{2 \cdot c_y}{b \cdot s}.
\end{aligned}\end{equation}

The orthogonal projection in the weak-perspective projection omits the $Z^{'}$ coordinate discrepancy inside the human body, which assumes the human body is far from the camera whose focal length is unrealistically large (corresponding to a very small field-of-view) \cite{kanazawa2018end,kolotouros2019learning}.
This is not true for many cases.
Thus we use the perspective projection with an appropriate focal length to calculate the 2D reprojection loss, because this is how the original image is captured.
However, we still let the model predict the weak-perspective projection parameters, since for most cases, $t_x \in [-1, 1]$, $t_y \in [-1, 1]$, $s \in [0, 1]$, meaning that they have the normalization property, which makes them suitable to be the CNN predictions.


\begin{figure}[t]
	\centering
	\includegraphics[width=\linewidth]{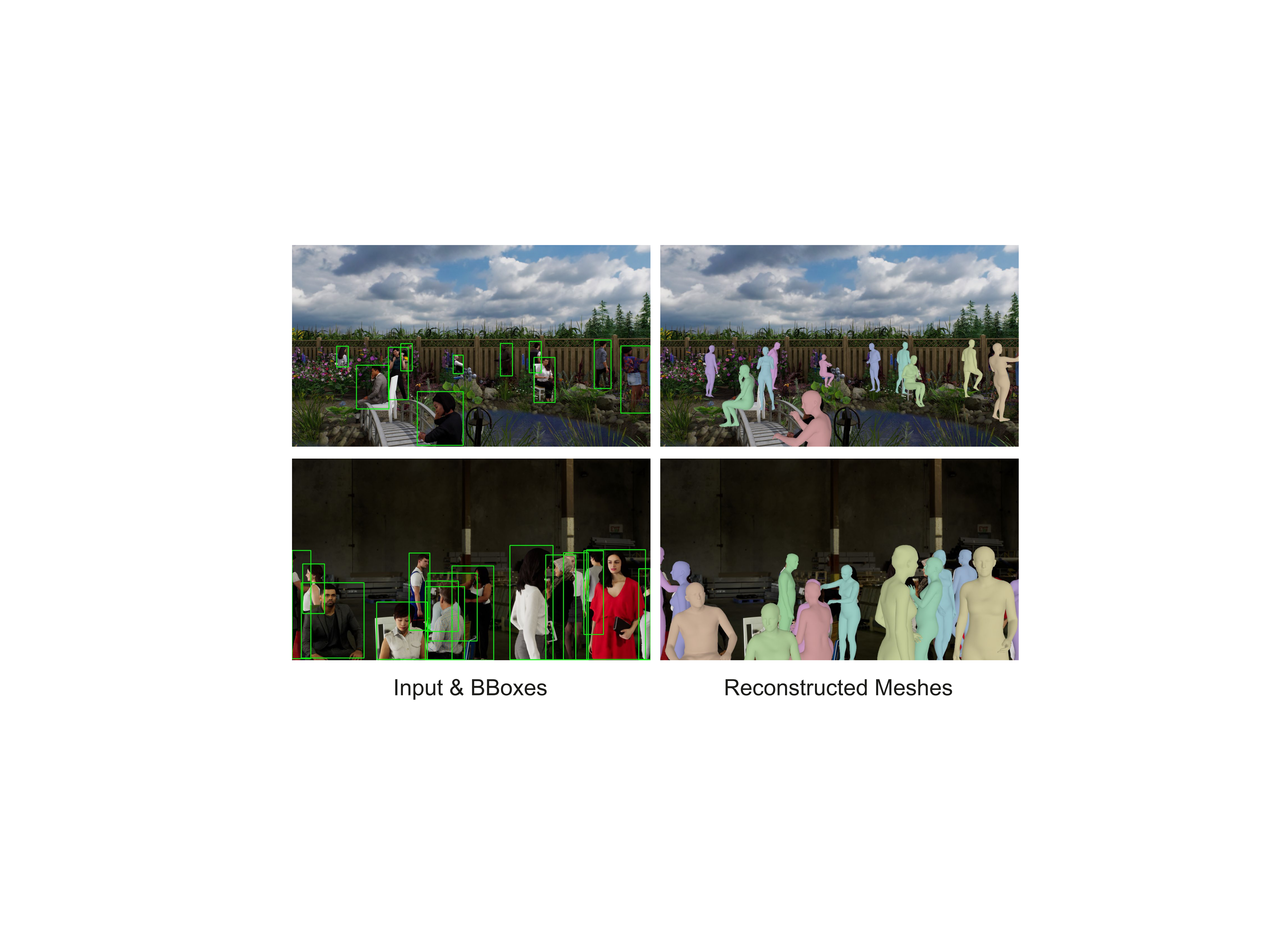}
	\caption {Qualitative results on the AGORA test samples. CLIFF works well in crowded and occluded scenes, even when large body parts are missing in the BBoxes.}
	\label{fig:agora}
\end{figure}

\section{Impact of the BBox Quality}
The BBox quality is important to our method, just like other top-down methods.
However, taking the BBox information as the additional input does not make our method rely more on the BBox quality.
As demonstrated in the AGORA evaluation, we use the BBox predicted by Mask R-CNN which is trained on COCO without finetuning on AGORA;
yet CLIFF still reaches the first place on the leaderboard, outperforming other top-down and bottom-up methods by large margins.
Note that AGORA contains a lot of crowded and severely occluded scenes, as shown in Fig.~\ref{fig:agora}.
CLIFF is robust to inaccurate BBox detection, mainly thanks to the data augmentation such as random scaling and cropping.

\begin{figure}[t]
	\centering
	\includegraphics[width=0.7\linewidth]{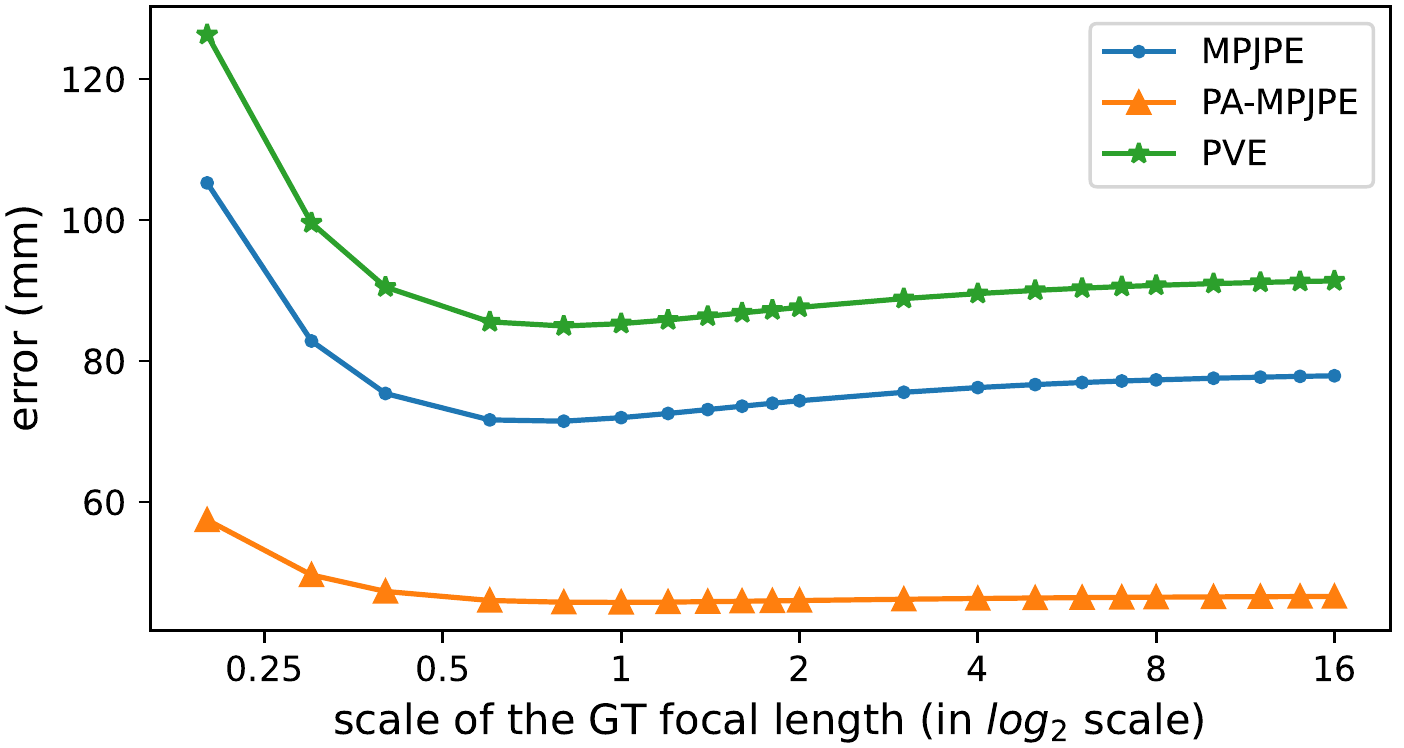}
	\caption {Impact of the focal length on estimation errors.}
	\label{fig:focal_l}
\end{figure}

\section{Impact of the Focal Length as Part of the Input}
We conduct this experiment on the 3DPW test set by perturbing the focal length from its GT value $f_{GT}$.
As shown in Fig.~\ref{fig:focal_l}, CLIFF is robust (with less than 5\% error increase) when the estimated focal length is in $[0.4f_{GT}, 3f_{GT}]$.
The estimation $f_{CLIFF} = \sqrt{w^2 + h^2}$ is within this range for most cases (except for super telephotos).
Moreover, in practical applications, $f_{GT}$ is often known, making the performance guaranteed.

\begin{table}[t]\scriptsize
	\centering
	\caption{ Comparison between CLIFF and video-based methods on 3DPW }
	\label{table:accel}
	\begin{tabular}{lcccc}
		\toprule
		Annotator & MPJPE $\downarrow$ & PA-MPJPE $\downarrow$  & PVE $\downarrow$ & Accel Error $\downarrow$ \\
		\midrule
		HMMR \cite{kanazawa2019learning}	& 116.5 & 72.6   & -    & 14.3    \\
		TCMR \cite{choi2021beyond}		& 86.5 & 52.7   & 102.9 & \B 7.1    \\
		VIBE \cite{kocabas2020vibe} 	& 82.7 & 51.9   & 99.1 & 23.4 \\
		MAED \cite{wan2021encoder} 		& 79.1 & 45.7   & 92.6 & 17.6 \\
		\midrule
		CLIFF (Res-50)					& 72.0 & 45.7 & 85.3 & 24.7 \\
		CLIFF (Res-50) w/ OneEuro		& 74.0 & 46.2 & 87.6 & 10.6 \\
		CLIFF (HR-W48)					& \B 69.0 & \B 43.0 & \B 81.2 & 20.5 \\
		CLIFF (HR-W48) w/ OneEuro		& 70.1 & 43.1 & 82.3 & 11.3 \\
		\bottomrule
	\end{tabular}
\end{table}

\section{Smoothness Comparison with Video-Based Methods}
We can apply CLIFF to a video frame by frame, and perform temporal smoothing to reduce jitter, such as OneEuro filtering \cite{casiez20121}.
Video-based methods \cite{kanazawa2019learning,choi2021beyond,kocabas2020vibe,wan2021encoder} usually make temporally smooth 3D predictions, which is their advantage over frame-based methods.
However, they cost much computation by processing additional adjacent frames.
Here we compare CLIFF with these video-based methods, especially on the smoothness evaluation, as shown in Table \ref{table:accel}.
The metric for evaluating temporal smoothness is acceleration error, which measures the average difference between ground truth 3D acceleration and the predicted 3D acceleration of each joint in $mm/s^2$.
CLIFF, as a frame-based method, achieves comparable smoothness performance with video-based methods.
With the additional OneEuro filtering as post-processing which costs negligible extra computation, the smoothness performance is improved significantly with slightly larger pose errors, which are still much smaller than those of the competitors.

\section{CLIFF Annotator Training}

\begin{figure}[t]
	\begin{minipage}[t]{0.32\linewidth}
		\centering
		\includegraphics[width=\textwidth]{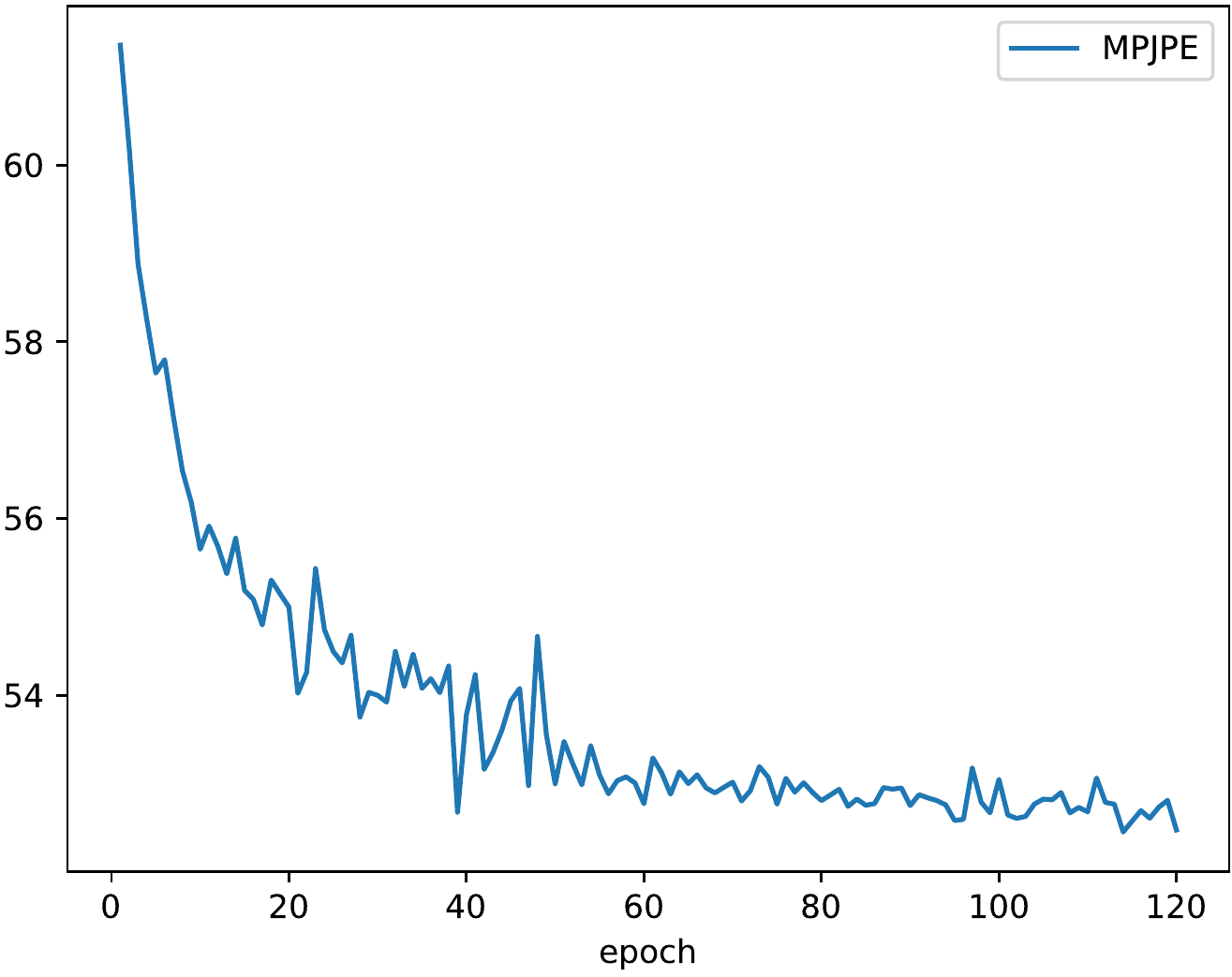}
	\end{minipage}
	\begin{minipage}[t]{0.32\linewidth}
		\centering
		\includegraphics[width=\linewidth]{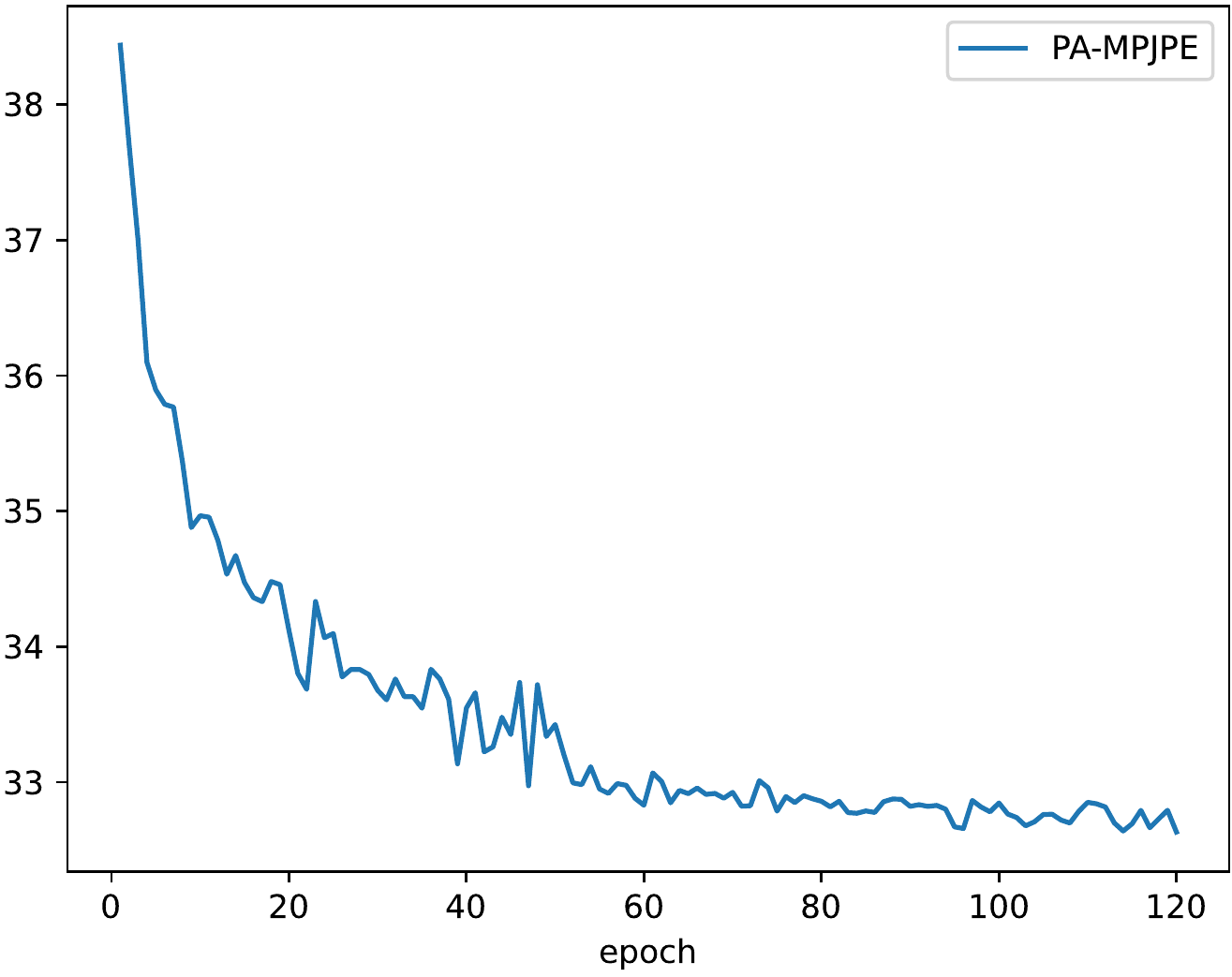}
	\end{minipage}
	\begin{minipage}[t]{0.32\linewidth}
		\centering
		\includegraphics[width=\linewidth]{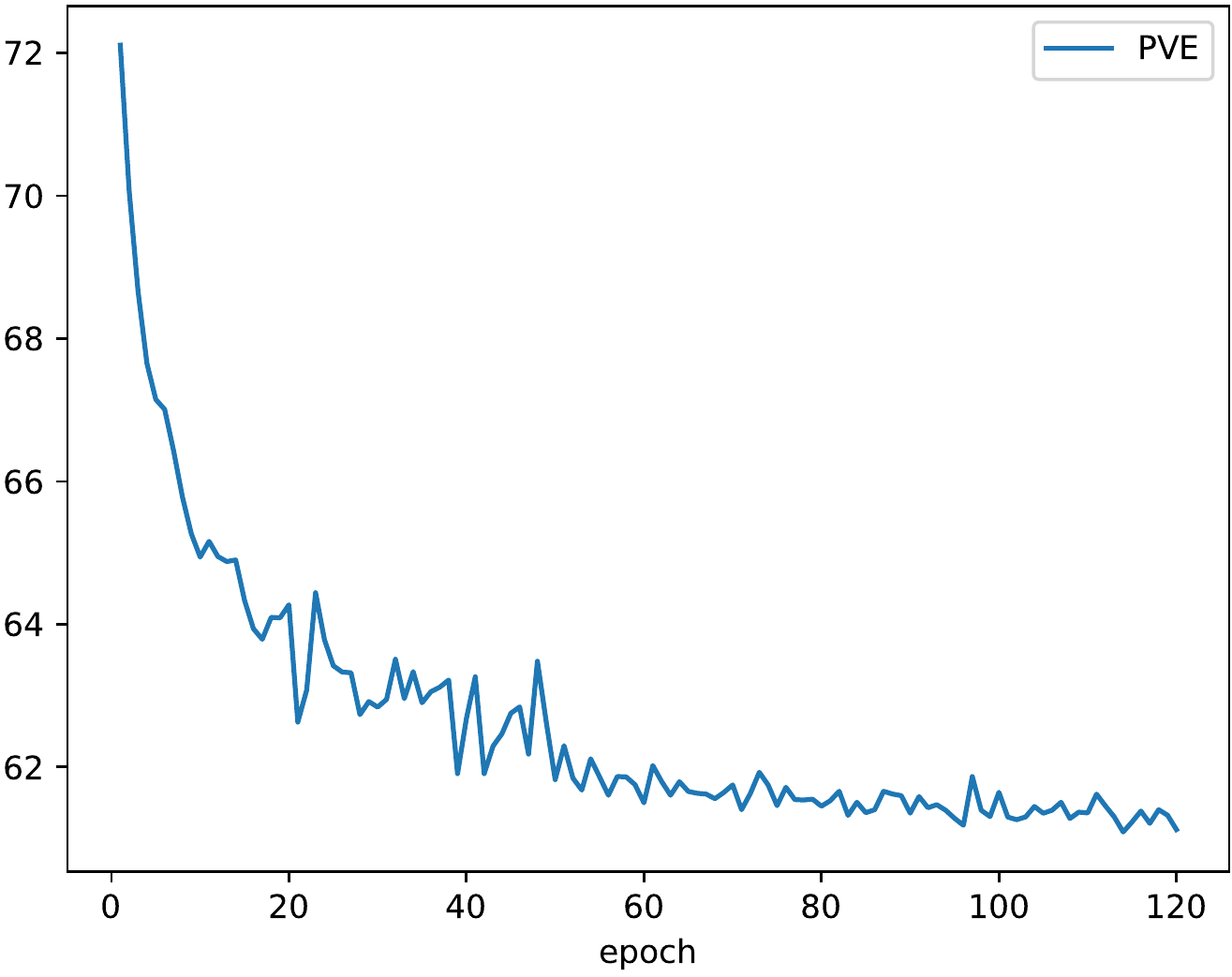}
	\end{minipage}
	\caption{The curves of training the CLIFF annotator on the 3DPW test data. The evaluation errors do not diverge even for long training, which means there is no need for the CLIFF annotator to choose a generic stopping criterion carefully.}
	\label{fig:annotaor_curve}
\end{figure}

In Fig. \ref{fig:annotaor_curve},  we show the evaluation error curves of training the CLIFF annotator on the 3DPW test data.
The learning rate starts from $5 \times e^{-5}$, and is reduced by a factor of 10 at the 45th epoch.
We can obtain a fine model before the 60th epoch, and the evaluation errors do not diverge even for longer training (120 epochs in total), and may decrease for a better performance.
It means that the CLIFF annotator is robust in the optimization, because the proposed priors prevent the annotator from overfitting to the 2D keypoints and from producing implausible poses.
Consequently, there is no need for our annotator to choose a generic stopping criterion carefully, which is a serious problem for EFT \cite{joo2021exemplar}.

\section{Ablation Study of the CLIFF Annotator}

\begin{table}[t]\small
	\centering
	\caption{ Ablation study of the CLIFF annotator on 3DPW }
	\label{table:cliff-annotator}
	\begin{tabular}{lccc}
		\toprule
		Annotator & MPJPE $\downarrow$ & PA-MPJPE $\downarrow$  & PVE $\downarrow$ \\
		\midrule
		ProHMR				& - & 52.4 & - \\
		BOA					& 77.2 & 49.5 & - \\
		EFT					& - & 49.3 & - \\
		DynaBOA				& 65.5 & 40.4 & 82.0 \\
		Pose2Mesh			& 65.1 & 34.6 & - \\
		\midrule
		HMR-based (Ours)  	& 63.6 & 38.7 & 72.6 \\
		{\bf CLIFF-based (Ours)} 	& \BB 52.8 & \BB 32.8 & \BB 61.5 \\
		\bottomrule
	\end{tabular}
\end{table}

We implement the proposed pseudo-GT annotator based on HMR, and compare it to the CLIFF-based one on 3DPW.
As shown in Table~\ref{table:cliff-annotator}, the errors increase when switching the base model from CLIFF to HMR, but the HMR-based annotator is still better than other SOTA methods.
Note that Pose2Mesh \cite{choi2020pose2mesh}, as a model-free method, produces only 3D vertices but no SMPL parameters.

\begin{figure}[t]
	\centering
	\includegraphics[width=\textwidth]{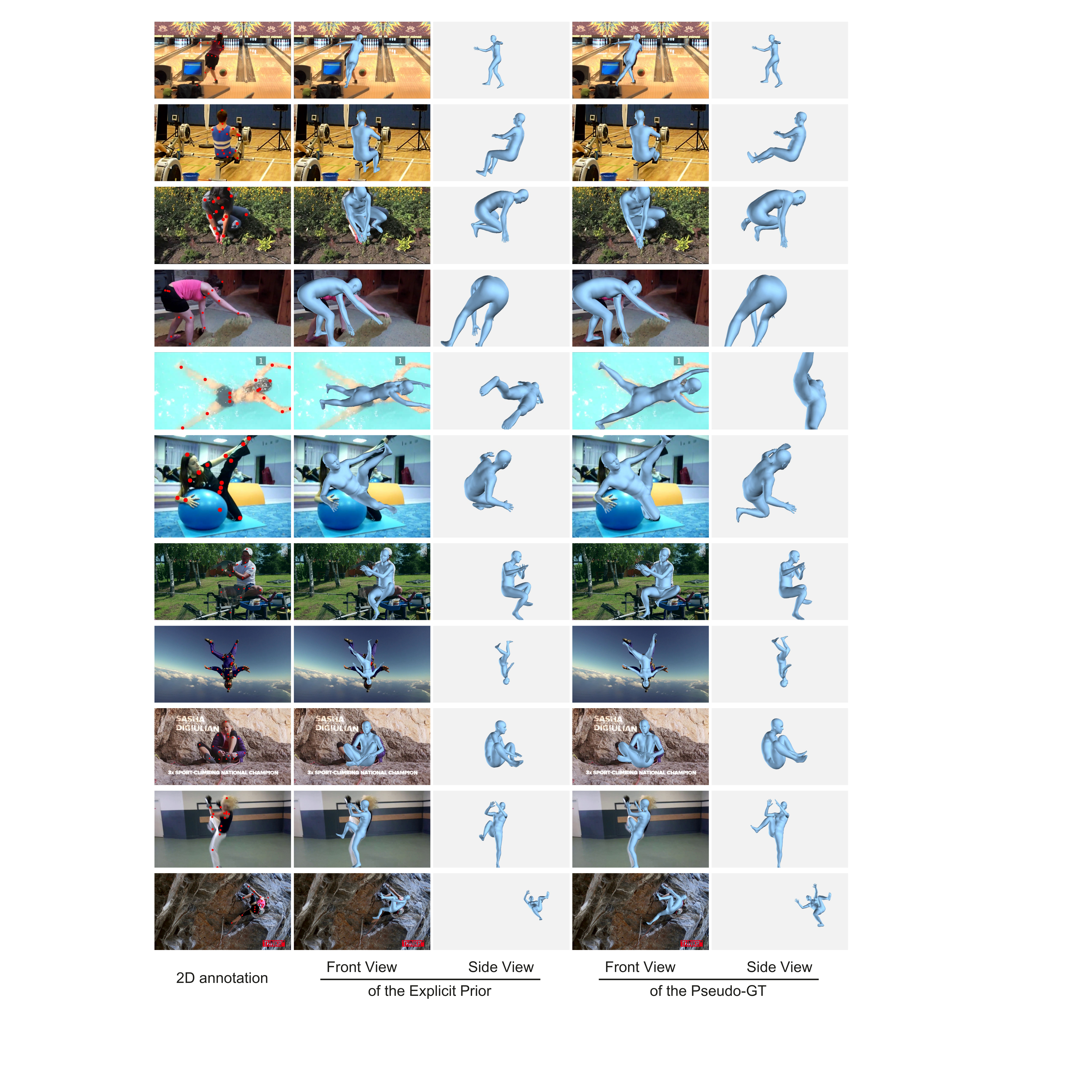}
	\caption{More pseudo-GT samples from the CLIFF annotator. From left to right: 2D annotation, front view of the explicit prior, side view of the explicit prior, front view of the pseudo-GT, and side view of the pseudo-GT.}
	\label{fig:pgt}
\end{figure}

\section{More Qualitative Pseudo-GT Results}
In Fig. \ref{fig:pgt}, we show additional qualitative results in the CLIFF annotator experiments. We test the pretrained annotator on the target images to get predictions as the explicit prior, which may not be accurate but usually plausible.
The final pseudo-GT achieves better pixel alignment, and maintains the plausibility with the help of the proposed priors.

\end{document}